%% file: main.tex
\documentclass[10pt,twocolumn,letterpaper]{article}

\usepackage[pagenumbers]{wacv} 

\usepackage{graphicx}
\usepackage{amsmath}
\usepackage{amssymb}
\usepackage{booktabs}
\usepackage{soul}
\usepackage[accsupp]{axessibility} 
\usepackage{url}

\usepackage[pagebackref,breaklinks,colorlinks]{hyperref}

\usepackage[capitalize]{cleveref}
\crefname{section}{Sec.}{Secs.}
\Crefname{section}{Section}{Sections}
\Crefname{table}{Table}{Tables}
\crefname{table}{Tab.}{Tabs.}

\def\wacvPaperID{8} 
\def\confName{WACV}
\def\confYear{2025}

\begin{document}

\title{Exploring Pose-Based Anomaly Detection for Retail Security: A Real-World Shoplifting Dataset and Benchmark}
\author{
  Narges Rashvand \quad Ghazal Alinezhad Noghre \quad Armin Danesh Pazho \quad Shanle Yao \quad Hamed Tabkhi\\
  University of North Carolina at Charlotte\\
  Charlotte, NC, USA\\
  {\tt\small \{nrashvan, galinezh, adaneshp, syao, htabkhiv\}@charlotte.edu}
}

\maketitle

\begin{abstract}
\input{Sections/Section_0_abstract}

\end{abstract}


\input{Sections/Secion_1_introduction}    
\input{Sections/Section_2_related_works}
\input{Sections/Section_3_dataset_process}

\input{Sections/Section_4_dataset_analysis}
\input{Sections/Section_5_metrics}

\input{Sections/Section_6_benchmarking}

\input{Sections/Section_7_conclusion}

{\small
\bibliographystyle{ieee_fullname}
\bibliography{egbib}
}

\end{document}

%% file: Sections/Section_0_abstract.tex
Shoplifting poses a significant challenge for retailers, resulting in billions of dollars in annual losses. Traditional security measures often fall short, highlighting the need for intelligent solutions capable of detecting shoplifting behaviors in real time. This paper frames shoplifting detection as an anomaly detection problem, focusing on the identification of deviations from typical shopping patterns. We introduce \textbf{PoseLift}, a privacy-preserving dataset specifically designed for shoplifting detection, addressing challenges such as data scarcity, privacy concerns, and model biases. PoseLift is built in collaboration with a retail store and contains anonymized human pose data from real-world scenarios. By preserving essential behavioral information while anonymizing identities, PoseLift balances privacy and utility. We benchmark state-of-the-art pose-based anomaly detection models on this dataset, evaluating performance using a comprehensive set of metrics. Our results demonstrate that pose-based approaches achieve high detection accuracy while effectively addressing privacy and bias concerns inherent in traditional methods. As one of the first datasets capturing real-world shoplifting behaviors, PoseLift offers researchers a valuable tool to advance computer vision ethically and will be publicly available to foster innovation and collaboration. The dataset is available at https://github.com/TeCSAR-UNCC/PoseLift.

%% file: Sections/Secion_1_introduction.tex
\section{Introduction}
\label{sec:intro}

\begin{figure}[t]
  \centering
   \includegraphics[trim=0pt 0pt 0pt 0pt, clip, width=3.4 in]{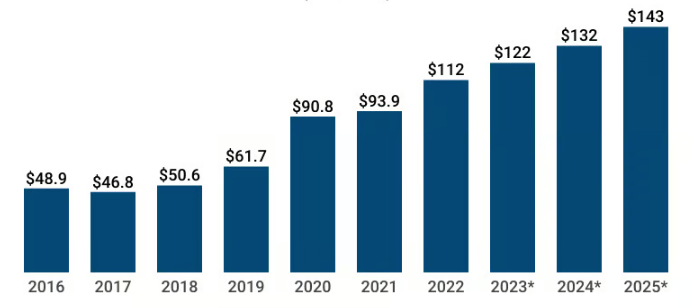}
   \caption{Retail revenue lost to shoplifting in the USA (in billions), with losses reaching nearly \$122 billion in 2023 and projected to surpass \$143 billion by 2025 \cite{shoplifting_capitalone2024, national_retail_federation}.}
   \label{fig:shoplifting_statistics}
\end{figure}

Shoplifting is a persistent issue that significantly impacts businesses, communities, and the economy. Retailers face substantial financial losses, operational inefficiencies, and security challenges due to undetected thefts \cite{ansari2022expert}. In the United States alone, retail theft resulted in \$112.1 billion in lost revenue in 2022 and an estimated \$121.1 billion in 2023. Projections indicate these losses could exceed \$143 billion by 2025 \cite{shoplifting_capitalone2024}, as shown in \cref{fig:shoplifting_statistics}. Despite these escalating figures, current security measures remain insufficient, with only about 2\% of shoplifters apprehended \cite{shoplifting_capitalone2024}.

The limitations of traditional security measures have driven significant interest in artificial intelligence-powered systems for shoplifting detection \cite{ansari2023optimized,kirichenko2022video}. Video surveillance systems, while ubiquitous, generate vast amounts of data that security personnel cannot analyze in real time, creating opportunities for automated detection systems \cite{arroyo2015expert, muneer2023shoplifting, pazho2023ancilia}. AI systems integrated with existing security infrastructures have the potential to enable real-time theft detection, generate alerts for personnel, and provide actionable insights, such as identifying high-risk areas and peak shoplifting times \cite{kirichenko2022detection}.

Despite advances in AI-based computer vision, research in vision-based shoplifting detection faces three critical challenges. First, the lack of real-world datasets hampers progress in shoplifting detection research. Existing datasets \cite{arroyo2015expert, ansari2022expert, muneer2023shoplifting} often rely on staged scenarios with actors or data aggregated from online sources \cite{sultani2018real}, failing to capture the complexities of authentic shoplifting incidents. Such datasets lack the contextual nuances specific to individual retail environments, limiting their applicability in real-world settings.

Second, the complexity and scarcity of labeled data further hinder progress. Shoplifting events are rare, unpredictable, and difficult to capture in real-world environments, making supervised learning approaches reliant on labeled data less effective \cite{pazho2023survey}. Furthermore, shoplifting behaviors vary significantly due to differences in store layouts, product types, and customer interactions, complicating efforts to define universal patterns.

Third, privacy and bias concerns are critical issues when detecting shoplifting behaviors \cite{alinezhad2023understanding, hirschorn2023normalizing}. Using raw video data raises privacy concerns, as it may capture identifiable customer information. Additionally, detection models are at risk of introducing biases based on factors such as gender, race, or clothing, potentially leading to unfair outcomes \cite{noghre2024pheva, benthall2019racial, buet2022towards, nagpal2019deep}.

To address these challenges, this research introduces PoseLift, a novel, privacy-preserving dataset for shoplifting detection. PoseLift is the result of close collaboration with a local retail store, capturing real-world shoplifting incidents alongside normal shopping behaviors. To the best of our knowledge, it is among the first datasets to incorporate genuine shoplifting scenarios while prioritizing privacy and stakeholder engagement. The dataset includes anonymized data, such as bounding boxes, tracking IDs, and human pose representations, which preserve essential behavioral information while ensuring privacy.

In addition to the dataset, this paper frames shoplifting detection as an anomaly detection problem, a topic that has been extensively studied across various domains by numerous researchers \cite {koran2024unveiling, abshari2024llm, doroudian2022study, zibaeirad2024comprehensive, ho2023graph}. To address the scarcity of labeled data and the diversity of shoplifting behaviors, we focus on unsupervised anomaly detection, which has been widely applied across domains \cite{ yu2019unsupervised, borghesi2019anomaly, park2018anomaly} and presents a promising approach to identifying deviations from normal patterns \cite{10314785}. These methods eliminate the need for labeled examples and adapt to diverse environments, making them well-suited for real-world retail applications.

We conduct comprehensive benchmarking experiments using state-of-the-art anomaly detection models, including STG-NF \cite{hirschorn2023normalizing}, TSGAD \cite{noghre2024exploratory}, and GEPC \cite{markovitz2020graph}, evaluated with metrics such as Area Under the Receiver Operating Characteristic Curve (AUC-ROC), Area Under the Precision-Recall Curve (AUC-PR), and Equal Error Rate (EER). The findings highlight the utility of pose-based anomaly detection for shoplifting detection and its potential to address privacy and bias concerns while improving detection accuracy.

The key contributions of this study are:
\begin{itemize} \item PoseLift: A privacy-preserving, real-world dataset with de-identified annotations for normal shopping and shoplifting behaviors.
\item Framing shoplifting detection as anomaly detection: Redefining shoplifting detection as an anomaly detection problem to identify atypical behaviors in retail environments.
\item Benchmarking anomaly detection models: Evaluating state-of-the-art pose-based anomaly detection models on PoseLift using comprehensive performance metrics.
\end{itemize}

This study lays the foundation for future research in privacy-preserving retail security, advancing the state-of-the-art in shoplifting detection systems. PoseLift is publicly available to support ongoing research and development in this critical and socially relevant domain.

%% file: Sections/Section_2_related_works.tex
\section{Related Works}

In the field of shoplifting detection, some studies focus on creating datasets, while others concentrate on developing algorithms for detection.
%

Muneer et al. \cite {muneer2023shoplifting} presents a large dataset with 900 videos. While the developed dataset is extensive, it lacks critical features for real-world applications. Notably, despite being labeled as a real-world dataset, it involves prearranged scenarios where individuals are asked to steal items in four predefined ways-placing items in their pockets, shirts, jackets, or college bags. Furthermore, these actions are performed exclusively by boys, which limits the diversity of the dataset. Additionally, all videos in this dataset were captured from a single camera angle, reducing the variety of perspectives needed for robust detection. 

In another dataset presented in \cite{ansari2022expert}, a shoplifting dataset was created using a 32 Megapixel camera. The proposed dataset consist of two classes: the normal class and shoplifting class. The normal class includes video clips depicting activities such as walking or inspecting shop items, while 
the shoplifting class contains clips of theft-related actions, such as concealing items under clothing or in bags. However, all videos were captured from a single camera angle, and the environment used is not a real retail store. 

In contrast, many studies on shoplifting detection, such as those by Kirichenko et al. \cite {kirichenko2022detection}, Nazir et al. \cite {nazir2023suspicious}, and Ansari et al \cite {ansari2023optimized}, have introduced algorithms that detect shoplifting at the pixel level. These studies rely on the UCF-Crime dataset \cite {sultani2018real}, the only publicly available dataset containing real-time shoplifting video clips. For instance, 
Kirichenko et al. \cite {kirichenko2022detection} addressed the problem of shoplifting by proposing a hybrid neural network to classify video fragments as either shoplifting or non-shoplifting, with their experiments conducted using the UCF-Crime dataset \cite {sultani2018real}.
Nazir et al. \cite {nazir2023suspicious} proposed a method for detecting suspicious behavior based on temporal feature extraction and time-series classification, using UCF-Crime dataset \cite {sultani2018real} for their experiments. Similarly, Ansari et al. \cite {ansari2023optimized} introduced a shoplifting detection system that combines YOLOv5 object detection \cite {yolov5} and Deep Sort tracking \cite {wojke2017simple}, also tested on the UCF-Crime dataset \cite {sultani2018real}. An automatic shoplifting detection system using surveillance videos was also presented in \cite{gim2020automatic}.
Their proposed system utilizes a Region of Interest (ROI) optical-flow fusion network to enhance feature detection and improve the accuracy of shoplifting behavior recognition \cite{gim2020automatic}. All of the experiments are also based on the UCF-Crime dataset \cite{sultani2018real}.
This dataset includes 1900 long, unedited videos of real-life criminal events across 13 categories, including violent accidents, fights, burglaries, explosions, arrests, arson, assault, traffic accidents, robberies, shootings, thefts, shoplifting, and vandalism \cite{sultani2018real}. The UCF-Crime videos are sourced from existing videos, all collected from online sources. While comprehensive, this dataset was not specifically designed for shoplifting detection. 

To overcome the limitations of existing datasets, including a single camera view, a limited sample size of shoplifting, and the use of prearranged shoplifting by actors, we developed a dataset specifically designed for shoplifting detection, featuring authentic real-world samples captured in actual store environments. This dataset incorporates various camera views and carefully addresses ethical considerations. To the best of our knowledge, it is among the first shoplifting datasets specifically tailored for pose-based shoplifting detection models.

%% file: Sections/Section_3_dataset_process.tex
\section{PoseLift Dataset}

This section provides an overview of the process involved in developing the dataset, covering data collection, preparation, and annotation, along with several segmented samples.

\subsection{Data Collection}
\begin{figure}[htt]
  
  \centering
   \includegraphics[trim=560pt 330pt 140pt 550pt, clip, width=3.3 in]{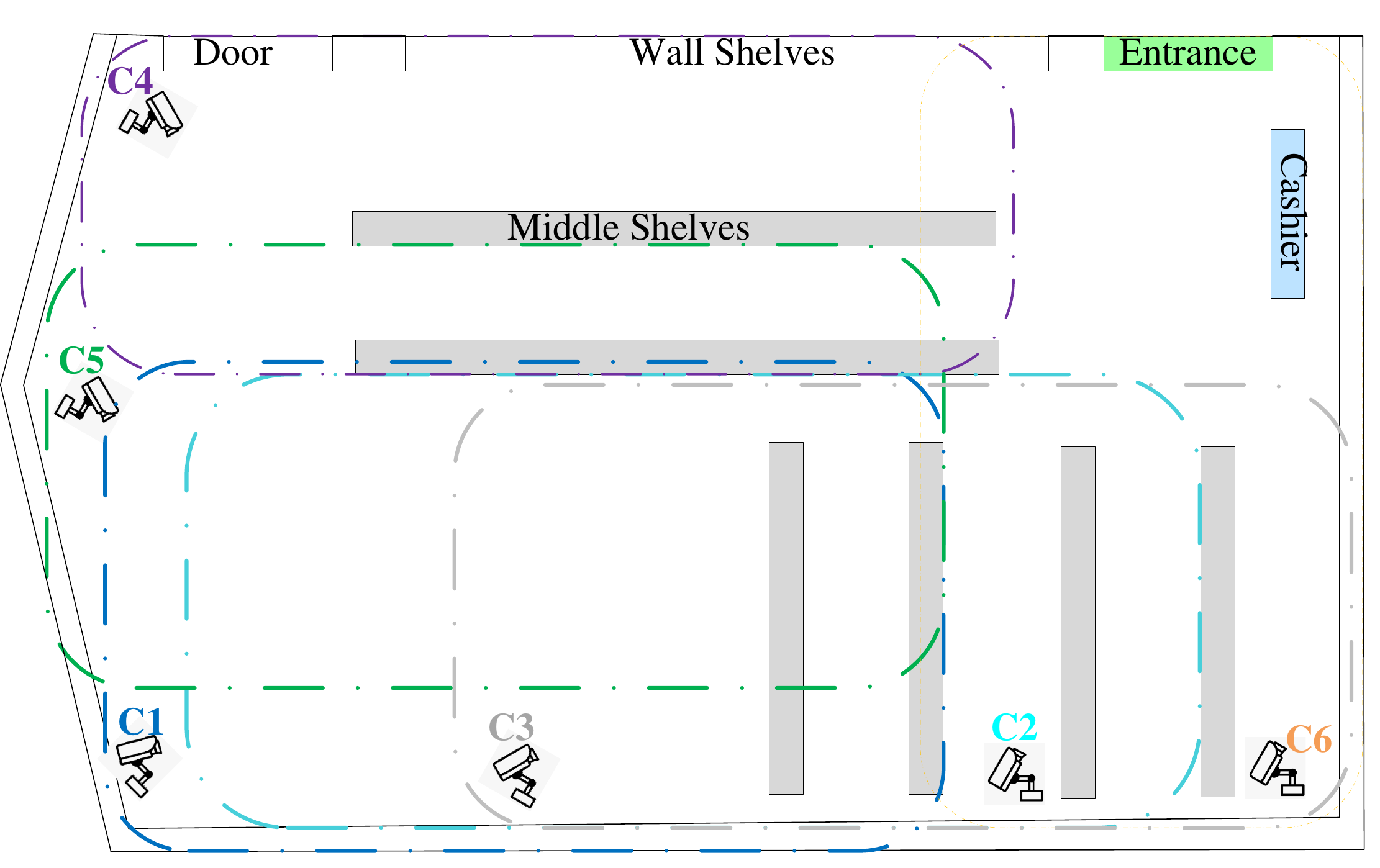}
   \caption{Bird's-eye view of the retail store, illustrating the locations of six cameras and their coverage area.}
   \label{fig:camera_locations}
\end{figure}

\begin{figure}[t]
  \centering
   \includegraphics[trim=570pt 90pt 570pt 85pt, clip, width=3.3 in]{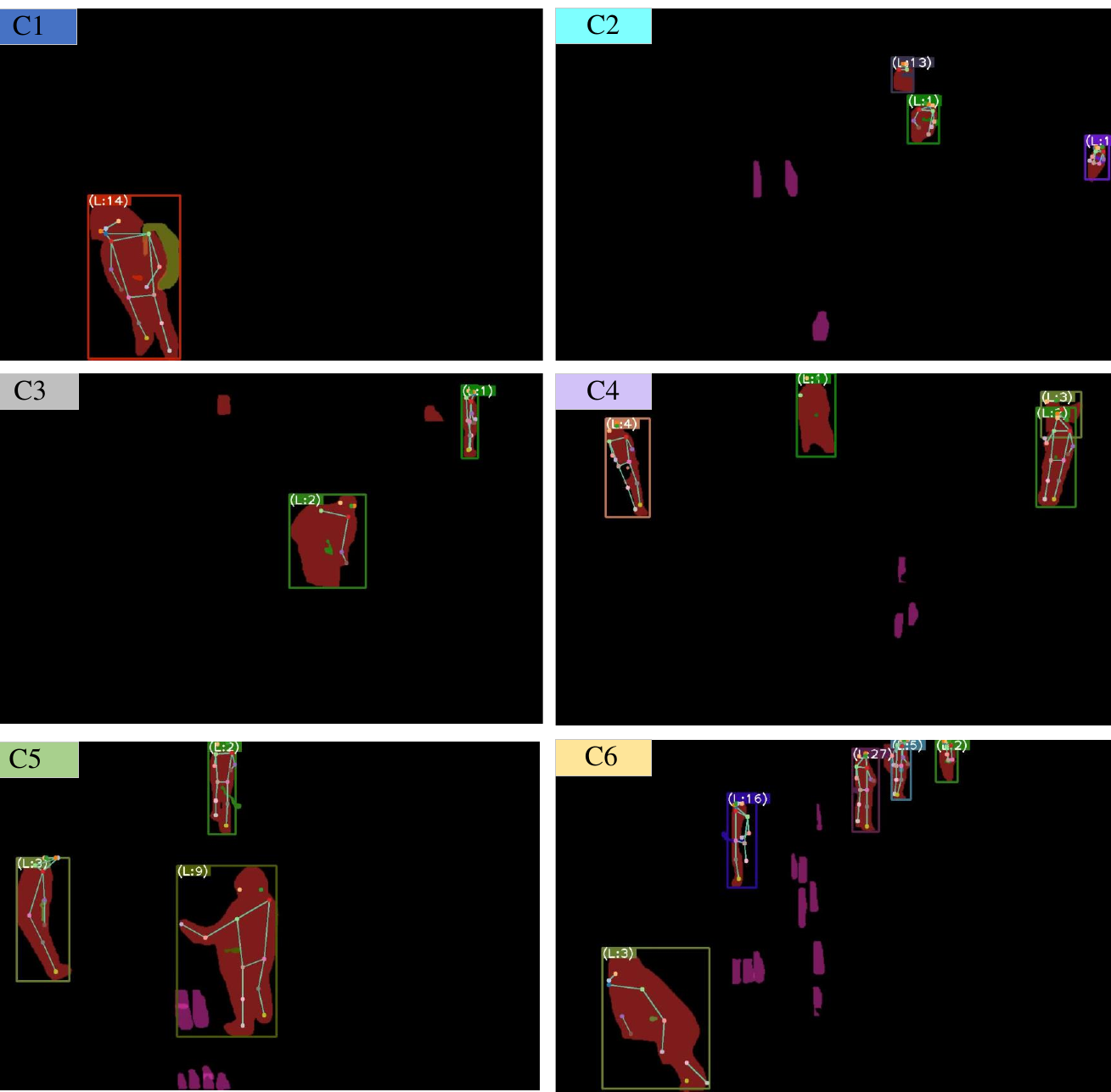}
   \caption{Segmented images from six camera views within a retail store, showcasing various perspectives used to capture normal and shoplifting instances in the dataset.}
   \label{fig:camera_viwes}
   \vspace{-10pt}
\end{figure}

The dataset was created using real-world data directly sourced from actual retail environments. We collaborated with stakeholders to obtain CCTV footage from a retail store, capturing real-world instances of both normal and shoplifting behaviors. The footage was recorded at a local retail store in the USA.
All the videos were captured from high-angle views with a resolution of  $1920\times1080$ and a frame rate of 15 frames per second. The original videos were processed to separate instances of shoplifting, resulting in a final collection of 155 videos. These videos vary in length and camera views, with the longest video being 331 seconds. 
The store is equipped with a total of 6 indoor cameras (C1 to C6), placed across various aisles and shelves in different locations throughout the store. The camera locations and their coverage are shown in the bird's-eye view in \cref {fig:camera_locations}.
Additionally,
\cref{fig:camera_viwes} illustrates the segmented camera views used in the dataset to capture normal and shoplifting video footage.

The distribution of videos by camera is as follows: 42 videos from C1, 33 videos from C2, 34 videos from C3, 21 videos from C4, 16 videos from C5, and 9 videos from C6. Among these, 43 videos contain instances of shoplifting. These shoplifting instances occur across different locations within the store, with varied camera views, providing a diverse range of theft activities. The shoplifting behaviors demonstrated in these videos included actions such as placing items into pockets, placing them in bags, and hiding them under shirts, jackets, and pants. This broad range of shoplifting behaviors ensures that the dataset effectively captures diverse shoplifting behaviors that can occur in real-world retail environments.

\subsection{Data Preparation, and Annotation}
As previously mentioned, the PoseLift dataset provides pose sequence data, specifically designed for pose-based shoplifting detection. To protect individual privacy, privacy-preserving techniques were applied by removing personal identifiers and anonymizing sensitive data. Instead of providing raw pixel-level video data, PoseLift offers pose sequence data, which captures abstract representations of human-body movements to ensure privacy and bias concerns mentioned in \cref{sec:intro}. 

In line with the principles of unsupervised anomaly detection and established methodologies in the field \cite{liu2018ano_pred, danesh2023chad, noghre2024pheva}, we have provided frame-level anomaly annotations for the PoseLift dataset. Trained annotators meticulously reviewed the footage, labeling each frame based on the presence or absence of shoplifting activities. This rigorous process involved multiple reviews of each video to ensure labeling accuracy. Shoplifting behaviors identified include actions such as placing items into pockets or bags, or concealing them under clothing. Conversely, standard shopping activities were labeled as normal. Consequently, each frame was classified as either shoplifting (anomalous, labeled as 1) or non-shoplifting (normal, labeled as 0). This labeling protocol was consistently applied across videos from cameras C1 through C6. 

The anonymized data provided by the PoseLift dataset includes bounding box annotations, person ID annotations, and human pose annotations. We have used a similar approach to \cite{pazho2023ancilia} for extracting the annotations. YOLOv8 object detection model \cite{yolov8_ultralytics} was employed to detect and localize individuals within each frame of the video. YOLOv8 generates bounding boxes around detected persons, indicating their positions within the scene. These bounding box annotations serve as a foundational step for subsequent stages, including human pose estimation and tracking. 

For person ID annotations, we have utilized the ByteTrack \cite {zhang2022bytetrack} algorithm, enabling robust tracking of multiple individuals across frames. ByteTrack ensures the tracking of individuals, even in crowded scenes, allowing the system to assign IDs to each person. This temporal tracking capability is essential for detecting progressing anomalies, as it enables the detection of shoplifting behaviors that unfold over time.

For human pose annotations we leveraged HRNet \cite {sun2019deep}, a state-of-the-art model for human pose estimation. HRNet extracts 2D skeletons by identifying keypoints on the human body, providing a detailed representation of body movements. These keypoints are presented according to the COCO17 \cite {lin2014microsoft} keypoint format, which includes 17 distinct points that define human body pose. 

To further improve the quality of the data, several modifications were applied. First, due to the presence of many shelves in the store and the resulting occlusions, we defined specific areas of interest for each camera and excluded individuals outside these designated regions. This allowed us to focus on relevant actions for each camera and minimize distractions caused by areas with limited visibility. To improve the accuracy and continuity of pose annotations, we applied linear interpolation to fill in any missing poses between frames. Additionally, an 8-frame window was used for data smoothing, ensuring that body movements appear continuous while reducing potential noise or errors in pose tracking. These steps collectively ensure the dataset's pose annotations are both accurate and reliable, making it well-suited for training robust shoplifting detection models.

\subsection{Samples from PoseLift Dataset}

 Since PoseLift exclusively provides pose sequence data rather than raw pixel data, we can only present selected segmented images from the developed dataset, highlighting various behaviors associated with both shoplifting and normal shopping activities.

\begin{figure}[htt]
  \centering
   
   \includegraphics[trim=80pt 300pt 80pt 90pt, clip, width=3.4 in]{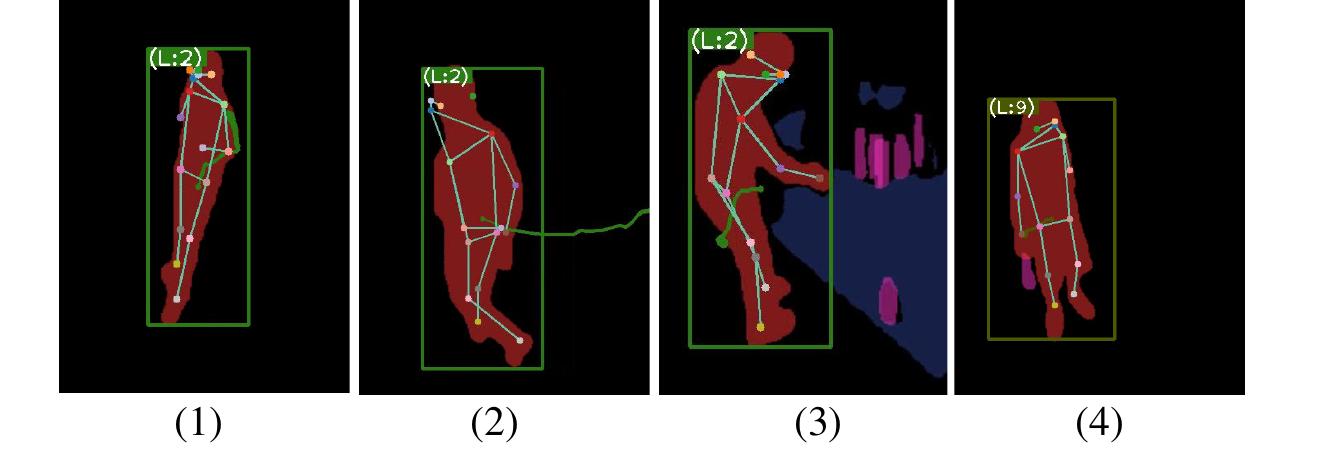}
   \caption{Four examples of normal shopping behavior, including browsing shelves (4.1), walking through the store (4.2), picking up an item (4.3), and carrying a bottle (4.4). These instances represent typical actions captured in PoseLift for comparison with shoplifting scenarios.}
   \label{fig:normal_samples}
\end{figure}

\cref {fig:normal_samples} presents four distinct examples of typical shopping behavior that are considered normal. As shown in the images, each image features 17 keypoints for each detected individual, along with a bounding box and an associated person ID. Fig. 4.1 depicts a person looking at items on the shelves. Fig. 4.2 shows an individual walking through the store, followed by Fig. 4.3, where an individual is picking an item from the shelf, and finally, Fig. 4.4 shows a person holding a bottle in their hands as they move through the store.

\vspace{-10pt}

\begin{figure}[htp]
  \centering
   \includegraphics[trim=75pt 450pt 150pt 470pt, clip, width=3.2 in]{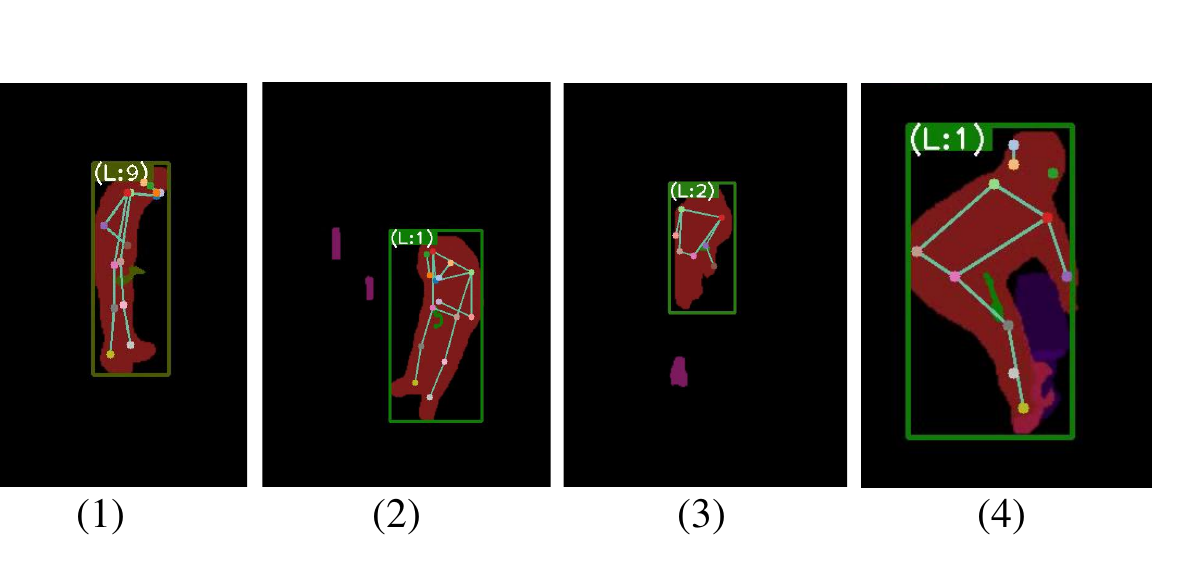}
   \caption{Four shoplifting instances captured from different angles, including concealing an item in their pants, placing an item in a front pocket, putting an item in an open bag, and placing an item in a bag on the ground. These instances illustrate typical shoplifting behaviors featured in our dataset.}
   \label{fig:shoplifting_samples}
\end{figure}

\cref{fig:shoplifting_samples} presents four instances of shoplifting captured in the PoseLift dataset. Fig. 5.1 shows an individual stealing an item and concealing it in their pants, with the scene viewed from a side angle. Building on this, Fig. 5.2 captures a similar scenario but from the front, where a person is seen secretly stealing an item and putting it into the front pocket of their T-shirt. In contrast, Fig. 5.3 illustrates a different technique, where an individual intentionally leaves their bag open and 
places a stolen item inside. Finally, Fig. 5.4 shows a shoplifter who steals an item and places it into a bag resting on the ground, with the view captured from behind. These images together highlight a variety of shoplifting behaviors across different perspectives.

\begin{figure*}[htp]
  \centering
   \includegraphics[trim=280pt 340pt 460pt 200pt, clip, width=6.6in]{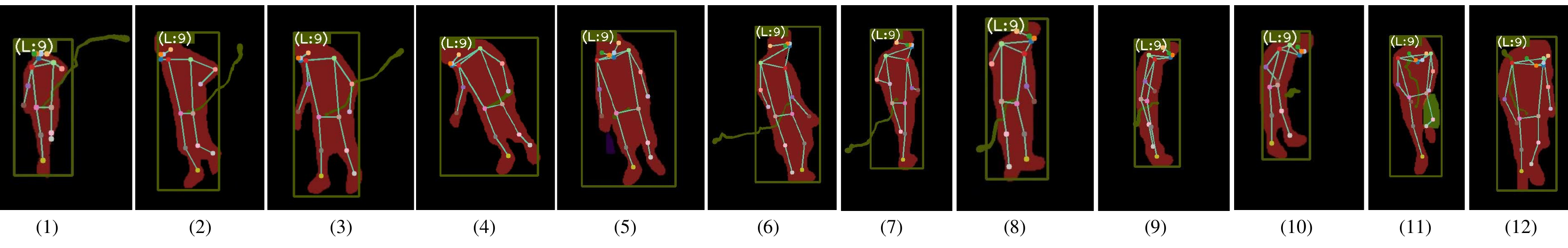}
   \caption{Pose sequence of a shoplifting incident across different frames, altering between normal and anomalous frames. The sequence shows the person approaching the shelf (Fig. 6.1-6.2), reaching for and grabbing an item (Fig. 6.3-6.4), carrying it (Fig. 6.5), adjusting their pants to conceal it (Fig. 6.7-6.10), and walking away (Fig. 6.11-6.12).}
   \label{fig:shoplifting_over_time}
\end{figure*}

In \cref{fig:shoplifting_over_time}, a single sample of shoplifting is presented across different frames, altering between normal and anomalous frames, to highlight the progression of the behavior over time. The pose sequence begins in Fig 6.1, where the person walks toward the shelf. In Fig 6.2, the person examines the shelves, and in Fig 6.3, they reach for an item. Fig 6.4 shows the person grabbing the item, while in Fig 6.5 the person is shown carrying the item. In Fig 6.6, the person rotates and moves to a different angle within the camera's view. In Fig 6.7, the person adjusts their pants to hide the item. Fig 6.8 and 6.9 show that person trying to hide the item in their pants, continuing in Figure 6.10. Finally, in Fig 6.11, the person begins to walk away, as shown in Fig 6.12.

%% file: Sections/Section_4_dataset_analysis.tex
\section{PoseLift Analysis and Comparison with Existing Datasets}

PoseLift contains 155 videos captured from 6 different camera views. Of these, 112 videos include normal behaviors, while 43 capture shoplifting behaviors. Since the dataset provides pose sequence data from real-world behaviors, we define a scenario based on the detection of a person until their person ID disappears. This allows us to differentiate between normal shopping scenarios and shoplifting scenarios. 
In total, the dataset contains 1,500 anomalous frames, captured across 43 shoplifting scenarios. In contrast, there are 779 normal scenarios from six different camera perspectives, which correspond to 55,574 frames of normal frames.

\cref {tab:dataset_comparision} presents a comparison between the PoseLift dataset and existing datasets in the field of shoplifting detection. As shown in the table, PoseLift is one of the first datasets sourced from real-world data captured within a single retail store. In contrast, datasets such as those by Arroyo et al. \cite{arroyo2015expert}, Ansari et al. \cite{ansari2022expert}, and Muneer et al. \cite{muneer2023shoplifting} rely on staged behaviors, typically involving actors. While the UCF-Crime dataset \cite{sultani2018real} includes real-world videos, it gathers videos from online sources like YouTube, sourced from multiple retail locations. What distinguishes PoseLift is its approach of collaborating with stakeholders to collect data directly from CCTV cameras in a retail store, ensuring the capture of normal and shoplifting behavior in a realistic retail setting.

Considering the length of the datasets and the distribution of shoplifting and normal videos, UCF-Crime \cite {sultani2018real} is the longest, with a total duration of 460,800 seconds. However, it is important to note that UCF-Crime \cite {sultani2018real} covers 13 different categories of anomalies and is not specifically designed for shoplifting detection. It includes 50 instances of shoplifting in different retail stores. 
In contrast, while PoseLift contains fewer shoplifting videos than datasets with prearranged shoplifting scenarios featuring actors, such as those by Arroyo et al. \cite {arroyo2015expert}, Ansari et al. \cite {ansari2022expert}, Iqra Muneer \cite {muneer2023shoplifting}, it offers realistic instances of shoplifting captured in actual retail environments. Additionally, compared to the shoplifting subset of the UCF-Crime  \cite {sultani2018real}, which contains collected real video footage, PoseLift contains a nearly identical number of shoplifting videos. Furthermore, in terms of total dataset length, PoseLift surpasses the length of those by Arroyo et al. \cite {arroyo2015expert}, Ansari et al. \cite {ansari2022expert}, Iqra Muneer \cite {muneer2023shoplifting}.

On the other hand, PoseLift is the only one that addresses privacy concerns by using pose sequence data, whereas all other datasets rely on raw pixel data. This privacy-preserving approach is the key distinction, and due to this difference, we are unable to directly compare the number of shoplifting and normal scenarios, as certain features are not accessible for comparison.

The datasets also vary in terms of Frames Per Second (FPS), with values of 10, 15, and 30 FPS. Furthermore, while other datasets are captured from a single camera view, with the exception of UCF-Crime \cite{sultani2018real}, PoseLift stands out as the only one that incorporates six distinct camera views. This diversity in camera perspectives enhances the robustness and applicability of PoseLift for real-world retail settings. 

\begin{table*}[htp]
  \centering
  \caption{Comparison of the PoseLift dataset, highlighting differences in the data source, data type, number of shoplifting and non-shoplifting videos, dataset length, FPS, and camera views.}
  \resizebox{\textwidth}{!}{
  \begin{tabular}{@{}c c c c c c c c c@{}}
    \toprule
    \toprule
    \textbf{Dataset}  &\textbf{Data Source} & \textbf{Type} & \textbf{\# Shoplifting Videos} & \textbf{\# Non-Shoplifting Videos} & \textbf{\# Total Videos} & \textbf{Length (s)} & \textbf{Avg FPS} & \textbf{\# Cameras} \\
    \midrule
    Arroyo et al. \cite{arroyo2015expert} & Staged & Raw & 155 & 755 & 910 & 2730 & 10  & 1  \\
    Ansari et al. \cite{ansari2022expert} & Staged & Raw & 87  & 88  & 175 & 1750 & 15  & 1  \\
    UCF-Crime \cite{sultani2018real} & Collected & Raw & 50  & 1850 & 1900 & 460,800 & 30  & -  \\
    Muneer et al. \cite{muneer2023shoplifting} & Staged & Raw & 450 & 450  & 900 & 2700 & 30  & 1  \\
    PoseLift & Real-World& Pose Seq & 43  & 112  & 155 & 3804 & 15  & 6  \\
    \bottomrule
    \bottomrule
  \end{tabular}}

  \label{tab:dataset_comparision}
\end{table*}

%% file: Sections/Section_5_metrics.tex
\section{Metrics}
In the context of shoplifting detection, it is essential to choose the appropriate evaluation metrics to assess model performance effectively. Given the class imbalance typically encountered in such tasks, where instances of shoplifting are much less than normal behaviors, metrics like AUC-ROC, AUC-PR, and EER provide complementary insights. These metrics will be discussed in the following subsections. 

\subsection{AUC-ROC}
AUC-ROC (Area Under the Receiver Operating Characteristic Curve) is a widely used performance metric in binary classification tasks, such as anomaly detection. The ROC curve visually represents a model's ability to discriminate between the normal and anomalous classes. It plots the True Positive Rate (TPR) against the False Positive Rate (FPR), showing how well the model distinguishes anomalies from normal instances at various classification thresholds. However, since AUC-ROC does not take the false Negative Rate (FNR) into account, it is important to consider additional metrics for a more comprehensive evaluation of the model's performance.
\subsection{AUC-PR}
AUC-PR (Area Under the Precision-Recall Curve) is a performance metric used for binary classifiers, particularly effective for imbalanced datasets. It evaluates a model's ability to identify the positive class by plotting precision against recall at various thresholds. A higher AUC-PR value indicates better performance in detecting anomalies.
\subsection{EER}
EER (Equal Error Rate) is a performance metric used in binary classification that identifies the point where the FPR equals the false negative rate (FNR), meaning the rate of false positives (normal instances incorrectly classified as shoplifting) matches the rate of false negatives (failing to detect actual shopliftings). A lower EER indicates better model performance, as it reflects a balanced trade-off between these two types of errors. In applications like shoplifting detection, where it is crucial to reduce both false positives and false negatives, EER proves to be an essential metric for evaluating model effectiveness.

%% file: Sections/Section_6_benchmarking.tex
\begin{figure*}[htp]
  \centering
   \includegraphics[trim=50pt 710pt 20pt 450pt, clip, width=4.0 in]{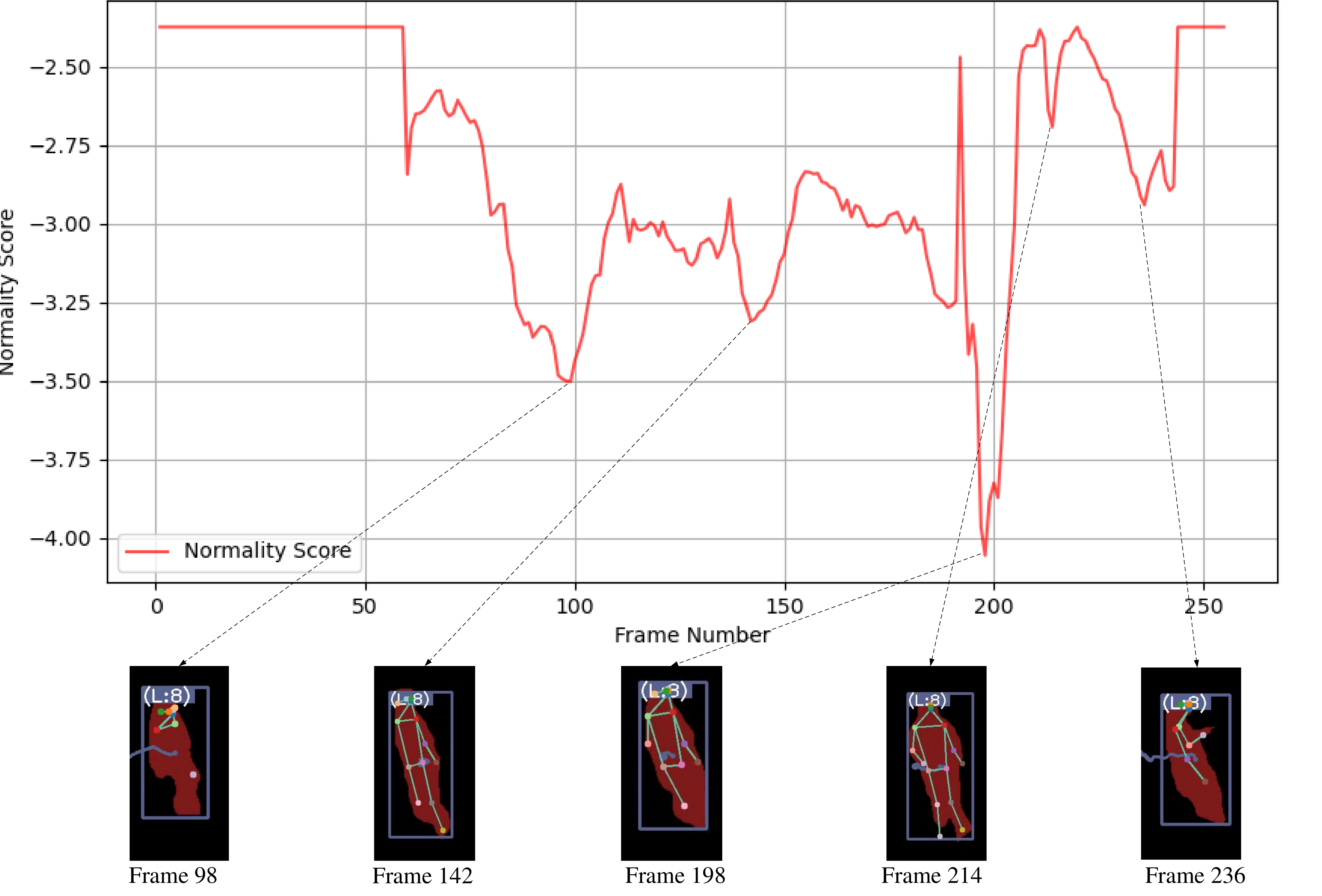}
   \caption{Normality score vs frame number plot for a false positive example using the STG-NF model on the PoseLift dataset. The plot shows low normality scores for normal activities (lower score translates to more anomalous), like walking and using a cell phone, which is mistakenly identified as shoplifting.}
   \label{fig:STG_NF_failure}
\end{figure*}

\begin{table*}[htp]
  \centering
  \caption{Separation of train and test sets for model training and evaluation, with nearly equal numbers of normal and anomalous frames. While the counts of normal and anomalous scenarios differ, this imbalance reflects the nature of shoplifting, which typically occurs more quickly than regular shopping behaviors. }
  \resizebox{\textwidth}{!}{
  \begin{tabular}{@{}c c c c c c c c@{}}
    \toprule
    \toprule
    \textbf{Dataset} & \textbf{\#Normal Frames} & \textbf{\#Anomalous Frames} & \textbf{\#Normal Scenarios} & \textbf{\#Shoplifting Scenarios} & \textbf{\#Camera Views}  \\
    \midrule
   Train Set & 53,353&  0& 760 & 0 &  6  \\
    Test set & 2,221 &  1,500 & 19 & 43 & 6  \\
    
    \bottomrule
    \bottomrule
  \end{tabular}}

  \label{tab:dataset_separation}
\end{table*}

\section{Model Benchmarking}
We employ available state-of-the-art pose-based anomaly detection models to demonstrate the effectiveness of the developed dataset. Specifically, we utilize STG-NF \cite{hirschorn2023normalizing}, GEPC \cite{markovitz2020graph}, and TSGAD \cite {noghre2024exploratory}, all of which are trained within an unsupervised learning framework. In this approach, the models are trained to minimize a specific objective that allows them to learn the normal behavior in retail environments. Once trained, the models are then evaluated on a separate test set to assess their ability to detect shoplifting incidents as anomalies.

The separation of training and test sets is crucial for providing an unbiased evaluation of model performance. Detailed information on this separation can be found in \cref{tab:dataset_separation}. 
As shown, the training set includes only normal shopping scenarios, allowing the models to learn typical customer behavior. The test set is carefully designed to maintain a balance between normal and anomalous instances, ensuring that the models are not biased toward one category. As shown in \cref{tab:dataset_separation}, the number of normal and anomalous frames are nearly identical, with 1,500 anomalous frames and 2,221 normal frames. Although the number of normal and abnormal scenarios differs, 19 normal scenarios compared to 43 shoplifting scenarios, this imbalance reflects the nature of the shoplifting scenarios, which tend to occur more rapidly than typical shopping behaviors. This training and test set division is used consistently for the training and evaluation of all models.

In the training phase, we used the default settings from the original papers for each model. For STG-NF \cite{hirschorn2023normalizing}, we set a batch size of 256 and trained the model for 3 epochs with a learning rate of 0.0005. For GEPC \cite{markovitz2020graph}, the model was trained with a batch size of 512, 10 epochs for the autoencoder, and 25 epochs for the deep clustering module. TSGAD \cite {noghre2024exploratory}, which employs a variational autoencoder and we focused only on the pose branch of this model, was trained with a batch size of 256, a  0.3 dropout rate, and 5 epochs with a learning rate of 0.00005. All models, except STG-NG, used the Adam optimizer, while STG-NF used the Adamx optimizer.

\cref{tab:metrics}  provides a detailed comparison of the models evaluated on the PoseLift dataset. Among them, STG-NF \cite{hirschorn2023normalizing} consistently outperforms the other models, achieving the highest AUC-ROC of 67.46\% and AUC-PR of 84.06\%  with the EER of 0.39. TSGAD \cite {noghre2024exploratory} ranks second in terms of AUC-ROC performance, while GEPC \cite{markovitz2020graph} has the lowest AUC-ROC of 60.61\% with its AUC-PR performance falling in the middle range. The superior performance of STG-NF can be attributed to its probabilistic design, which allows it to train effectively to capture the characteristics of normal behavior.

\begin{table}[htp]
\centering
\caption{Comparison of three performance metrics (AUC-ROC, AUC-PR, and EER) across three state-of-the-art models on the PoseLift dataset.}
\label{tab:metrics}
\resizebox{0.9\columnwidth}{!}{%
\begin{tabular}{lc|ccc}
\toprule[\heavyrulewidth] \midrule
\textbf{Methods}     & \textbf{Venue}  & \textbf{AUC-ROC} & \textbf{AUC-PR} & \textbf{EER} \\ \midrule
\textbf{STG-NF \cite{hirschorn2023normalizing}}   & ICCV 2023 & 67.46         & 84.06 & 0.39 \\
\textbf{TSGAD \cite{noghre2024exploratory}}   & WACV 2024 & 63.35      &  39.31  & 0.41 \\
\textbf{GEPC \cite{markovitz2020graph}}   & CVPR 2020 &  60.61  & 50.38   & 0.38  \\

\midrule \bottomrule[\heavyrulewidth]
\end{tabular}%
}
\end{table}

To highlight the challenges of the PoseLift dataset, we showcase an example of a false positive shoplifting detection. \cref {fig:STG_NF_failure} illustrates a failed case using the top-performing model, STG-NF \cite{hirschorn2023normalizing}, on the PoseLift dataset. The figure shows the model's normality score across multiple frames, all of which are normal. It is important to note that lower normality scores indicate greater deviation from what the model considers normal, implying a higher likelihood that the frame is an anomaly.

In frames 98 and 142, the person walks by a shelf and looks at it, with these frames having low normality scores (below -3.25). Additionally, in frames 198, 214, and 236, the person pulls out their cell phone from their pocket and uses it, which is assigned a low score and incorrectly flagged as shoplifting. Although anomaly detection models rely on these scores and thresholds to identify anomalies, setting different thresholds leads to false positives in this case. As a result, this example underscores the complexities of the PoseLift dataset, where the model's performance varies significantly across different instances. The observed variability emphasizes the necessity for more advanced, specialized models designed specifically for shoplifting detection, utilizing datasets engineered exclusively for this purpose.

%% file: Sections/Section_7_conclusion.tex
\section{Conclusion}

In this study, we introduced PoseLift, a benchmark dataset that offers pose sequences of real-world behaviors for shoplifting detection, addressing the limitations of previous approaches by focusing on privacy-preserving, real-world data. By framing shoplifting detection as an anomaly detection problem, we demonstrated the feasibility of using pose data to identify anomalous behaviors associated with shoplifting. The comprehensive evaluation of state-of-the-art anomaly detection models on PoseLift highlights the potential of this approach to mitigate privacy and bias concerns. The PoseLift dataset is publicly available to support future research and development, and we are continuously collecting more data to expand further and enhance the dataset.

\section{Acknowledgment}

This research is supported by the National Science Foundation (NSF) under Award Number 2329816.